\documentclass[letterpaper]{article} 
\usepackage{aaai2026}  
\usepackage{times}  
\usepackage{helvet}  
\usepackage{courier}  
\usepackage[hyphens]{url}  
\usepackage{graphicx} 
\usepackage{natbib}  
\usepackage{caption} 
\usepackage{algorithm}
\usepackage{algorithmic}

\usepackage{newfloat}
\usepackage{listings}
\DeclareCaptionStyle{ruled}{labelfont=normalfont,labelsep=colon,strut=off} 
\floatstyle{ruled}
\newfloat{listing}{tb}{lst}{}
\floatname{listing}{Listing}
\title{ForeDiffusion: Foresight-Conditioned Diffusion Policy via Future View Construction for Robot Manipulation}
\author{
    Weize Xie\textsuperscript{\rm 1, \equalcontrib}, 
    Yi Ding\textsuperscript{\rm 1, \equalcontrib}, 
		Ying He\textsuperscript{\rm 1, \thanks{Corresponding author.}}, 
    Leilei Wang\textsuperscript{\rm 1, \rm 2}, 
    Binwen Bai\textsuperscript{\rm 1}, 
    Zheyi Zhao\textsuperscript{\rm 1, \rm 2}, \\
    Chenyang Wang\textsuperscript{\rm 1}, 
    F. Richard Yu\textsuperscript{\rm 3} \\
}
\affiliations{
    \textsuperscript{\rm 1}College of Computer Science and Software Engineering, Shenzhen University, China \\
    \textsuperscript{\rm 2}Guangdong Laboratory of Artificial Intelligence and Digital Economy (SZ), Shenzhen, China \\
    \textsuperscript{\rm 3}School of Information Technology, Carleton University, Canada \\

    \{2410105060, 2410103031\}@mails.szu.edu.cn, heying@szu.edu.cn, wangleilei@gml.ac.cn, \\2300271042@email.szu.edu.cn, zhaozheyi@gml.ac.cn, chenyangwang@ieee.org
}

\usepackage{bibentry}

\usepackage{booktabs}
\usepackage{multirow}
\usepackage{array}
\usepackage{amsmath}

\begin{document}

\maketitle

\begin{abstract}
Diffusion strategies have advanced visual motor control by progressively denoising high-dimensional action sequences, providing a promising method for robot manipulation. However, as task complexity increases, the success rate of existing baseline models decreases considerably. Analysis indicates that current diffusion strategies are confronted with two limitations. First, these strategies only rely on short-term observations as conditions. Second, the training objective remains limited to a single denoising loss, which leads to error accumulation and causes grasping deviations. To address these limitations, this paper proposes Foresight-Conditioned Diffusion \textbf{(ForeDiffusion)}, by injecting the predicted future view representation into the diffusion process. As a result, the policy is guided to be forward-looking, enabling it to correct trajectory deviations. Following this design, ForeDiffusion employs a dual loss mechanism, combining the traditional denoising loss and the consistency loss of future observations, to achieve the unified optimization. Extensive evaluation on the Adroit suite and the MetaWorld benchmark demonstrates that ForeDiffusion achieves an average success rate of 80\% for the overall task, significantly outperforming the existing mainstream diffusion methods by 23\% in complex tasks, while maintaining more stable performance across the entire tasks.
\end{abstract}

\begin{links}
  \link{Code}{https://github.com/xwz-z/ForeDiffusion}
\end{links}

\begin{figure}[t]
    \centering
    \includegraphics[width=1.0\linewidth]{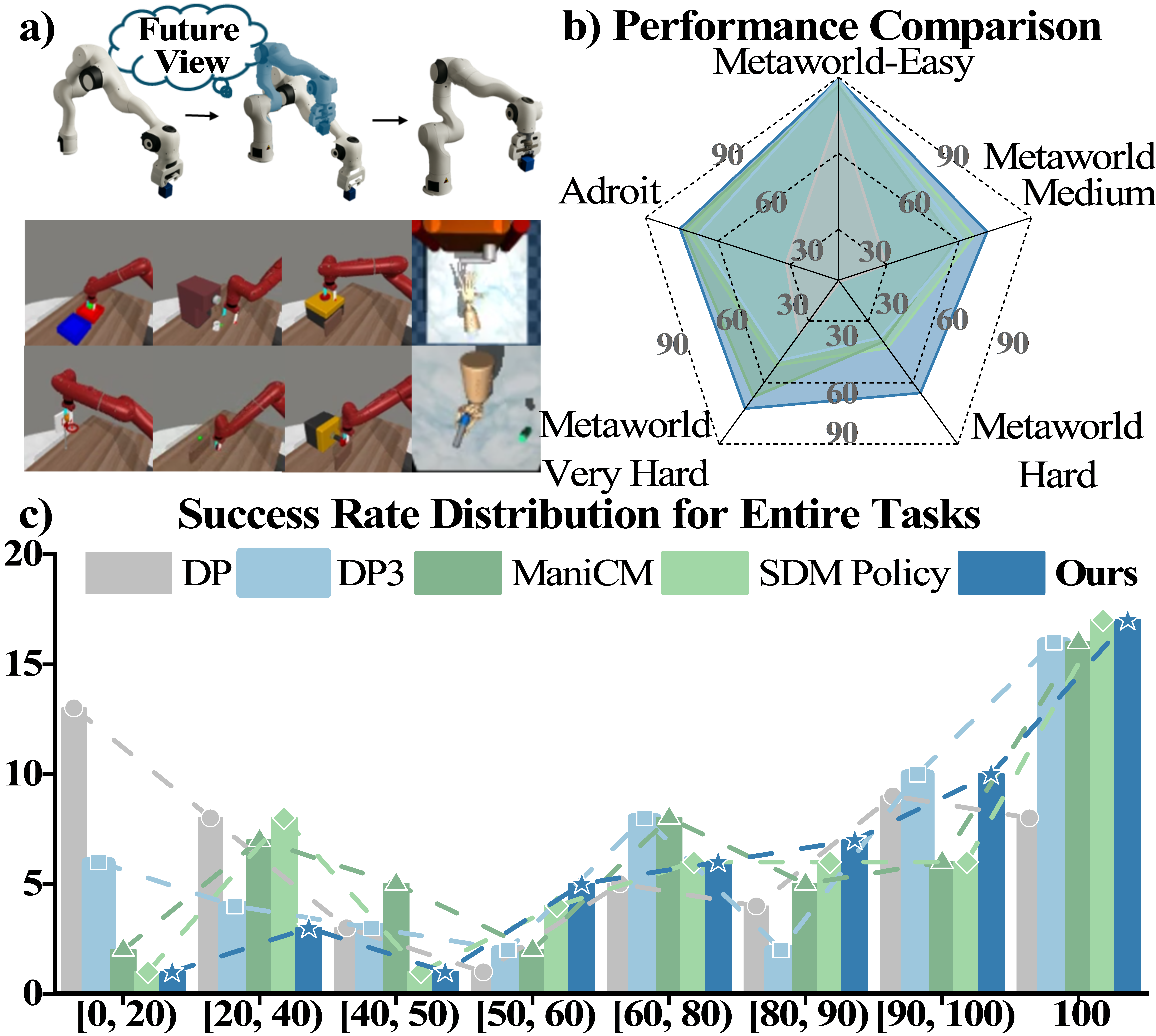}
    \caption{\textbf{Overview of ForeDiffusion.}
    (a) Diagram of future view-guided action generation; (b) ForeDiffusion achieves the highest average success rate across all task types; (c) ForeDiffusion shifts task counts toward the higher success rate bins across all tasks~\cite{chi2023diffusionpolicy, Ze2024DP3, lu2025manicmrealtime3ddiffusion,jia2024scoredistributionmatchingpolicy}.}
    \label{fig:Homepage}
\end{figure}
\section{Introduction}
Imitation learning from expert demonstrations offers a data-efficient supervised pathway to acquire diverse, task-conditioned manipulation competencies in embodied intelligence~\cite{rahmatizadeh2018vision,xie2020learning, zhao2024llakey}. Building on this paradigm, diffusion-based models have become expressive visual motion strategies for robotic tasks~\cite{urain2024deep}. In robotic manipulation, diffusion policies have been applied in increasingly rich perceptual settings to generate action sequences with feedback control~\cite{carvalho2023motion,wu2024tacdiffusion}. Early methods conditioned on RGB inputs have achieved reliable performance from monocular observations~\cite{chi2023diffusionpolicy}. The 3D Diffusion Policy shows that compact point cloud representations significantly enhance spatial understanding across multi-task benchmarks~\cite{Ze2024DP3}. Building on these advances, diffusion-based approaches have made substantial progress in multimodal conditioning by incorporating geometric, tactile, and proprioceptive signals~\cite{dou2024tactile,heng2025vitacformer, Cao2024MultiModalDF,zhao2025polytouch}.

However, current diffusion-based visuomotor policies still face significant limitations when applied to more complex manipulation tasks~\cite{wolf2025diffusion,xue2025reactive}. 
As shown in Fig.~\ref{fig:Homepage}b, mainstream diffusion models show significant performance degradation on complex tasks, which share the common characteristic of requiring sequential and contact-intensive interactions.~\cite{yu2020meta,lu2025manicmrealtime3ddiffusion}. Even small control inaccuracies can accumulate over time, resulting in task failure during later stages of execution~\cite{tsuji2025survey,stepputtis2022system}. 

We identify two primary factors causing this gap. First, current formulations typically condition diffusion on short observation sequences and assume approximate Markov sufficiency during rollout, without explicitly modeling how future scenes may evolve~\cite{reuss2023goal}. This limitation is especially problematic in tasks that span longer time horizons~\cite{kim2024stitching}. Second, most policies are trained with low-level losses like denoising or single step consistency, which focus on local accuracy but provide little guidance for completing full tasks~\cite{song2025survey,fan2025diffusion}. As a result, most fail with issues as misalignment, unstable grasps, or poor transitions~\cite{ma2024hierarchical}.

In this paper, we address limitations stemming from statically conditioned observation horizons and single training objectives. To overcome the challenges of existing diffusion-based methods in long-horizon, complex manipulation tasks, we propose \textbf{ForeDiffusion}, a foresight-conditioned diffusion policy. ForeDiffusion tackles these challenges by constructing a compact representation of future view and injecting it into the diffusion denoising process to guide policy rollout (Fig.~\ref{fig:Homepage}a). ForeDiffusion enables the model to reason beyond the immediate observation and mitigate cascading errors across multi-stage interactions. To support this mechanism, we introduce dual loss objective that combines standard denoising accuracy with a foresight‑consistency defined over the predicted horizon. Through this design, ForeDiffusion enhances planning capabilities while ensuring stability. The proposed method surpasses recent diffusion‑based baselines in overall tasks and high‑success task coverage across Adroit and MetaWorld (Fig.~\ref{fig:Homepage}b–c).
Our contributions are summarized as follows:
\begin{itemize}
    \item We introduce \textbf{ForeDiffusion}, which \textbf{injects a future view constructed} based on current observations into each denoising step, endowing diffusion policies with foresight, thus better addressing long-horizon tasks.
    \item We build on foresight-conditioned diffusion by formulating \textbf{dual loss objective} that combines standard denoising fidelity with prediction‑consistency, thereby curbing error accumulation in complex, contact‑rich manipulation.
    \item Compared with mainstream baselines, ForeDiffusion maintains superior overall performance and significantly  \textbf{improves the success rate by 23\% }in complex manipulation tasks, where other models perform poorly.
\end{itemize}

\section{Related Work}

\subsection{Visuomotor Control via Diffusion Models}
Diffusion models have emerged as a powerful paradigm for generating action sequences in robotic manipulation, leveraging iterative denoising processes inspired by their success in image and video synthesis~\cite{ho2020denoising,ho2022video}. Early works like Diffusion Policy are conditioned on RGB observations to achieve reliable performance in single-arm manipulation tasks, outperforming traditional behavior cloning methods~\cite{chi2023diffusionpolicy}. Recent advances incorporate richer perceptual inputs, such as 3D Diffusion Policy (DP3), which uses point cloud representations to enhance spatial reasoning and data efficiency on multi-task benchmarks like MetaWorld~\cite{Ze2024DP3}. Other innovations include FlowPolicy, which employs manifold-aligned denoising to capture low-dimensional action structures~\cite{zhang2025flowpolicy}, ManiCM integrates multimodal inputs into a consistency-driven 3D diffusion framework for real-time inference~\cite{lu2025manicmrealtime3ddiffusion}, and SDM Policy, which uses teacher-student distillation to enable single-step generation, improving task success rates~\cite{jia2024scoredistributionmatchingpolicy}. Recent surveys further categorize diffusion policies in grasp learning, trajectory planning, and skill acquisition~\cite{zhang2025unifying,ma2024hierarchical}. However, these approaches typically rely on static conditioning based primarily on initial proprioceptive inputs, without modeling temporal evolution, which limits foresight and leads to degraded performance in dynamic or complex tasks~\cite{lv2025spatial,reuss2023goal}. 

\subsection{Foresight-Driven Control with Dual Loss Training}

A parallel line of research studies predictive world models that forecast future observations, giving the agent foresight beyond its current sensory window~\cite{hafner2019learning}. Latent video predictor DreamerV3 rolls out imagined trajectories in a learned latent space to evaluate long-horizon returns~\cite{hafner2023mastering}. MoDem-V2 integrates RGB-based perception with learned visuo-motor models to perform contact-rich manipulation in uninstrumented, real-world environments~\cite{lancaster2024modem}. Hierarchical Diffusion Policy (HDP) and  Causal Diffusion Policy (CDP) split prediction across coarse and fine temporal scales, allowing long range forecasts to steer fine-grained actions~\cite{ma2024hierarchical,ma2025cdp}. Evidence from other fields confirms the benefit of explicit foresight.
SceneDiffuser and MotionDiffuser add cost or safety terms to diffusion based trajectory forecasts, lowering collision rates on Waymo and nuScenes~\cite{jiang2023motiondiffuser,jiang2024scenediffuser}. DiffuseLoco stabilizes quadruped gaits by penalizing high energy latent rollouts, while MVDiffusion and Percept-Diff pair pixel denoising with perceptual to keep long videos semantically coherent~\cite{huang2025diffuseloco,shi2023mvdream,borno2024percept}.

\begin{figure*}[t]
    \centering
    \includegraphics[width=1.0\linewidth]{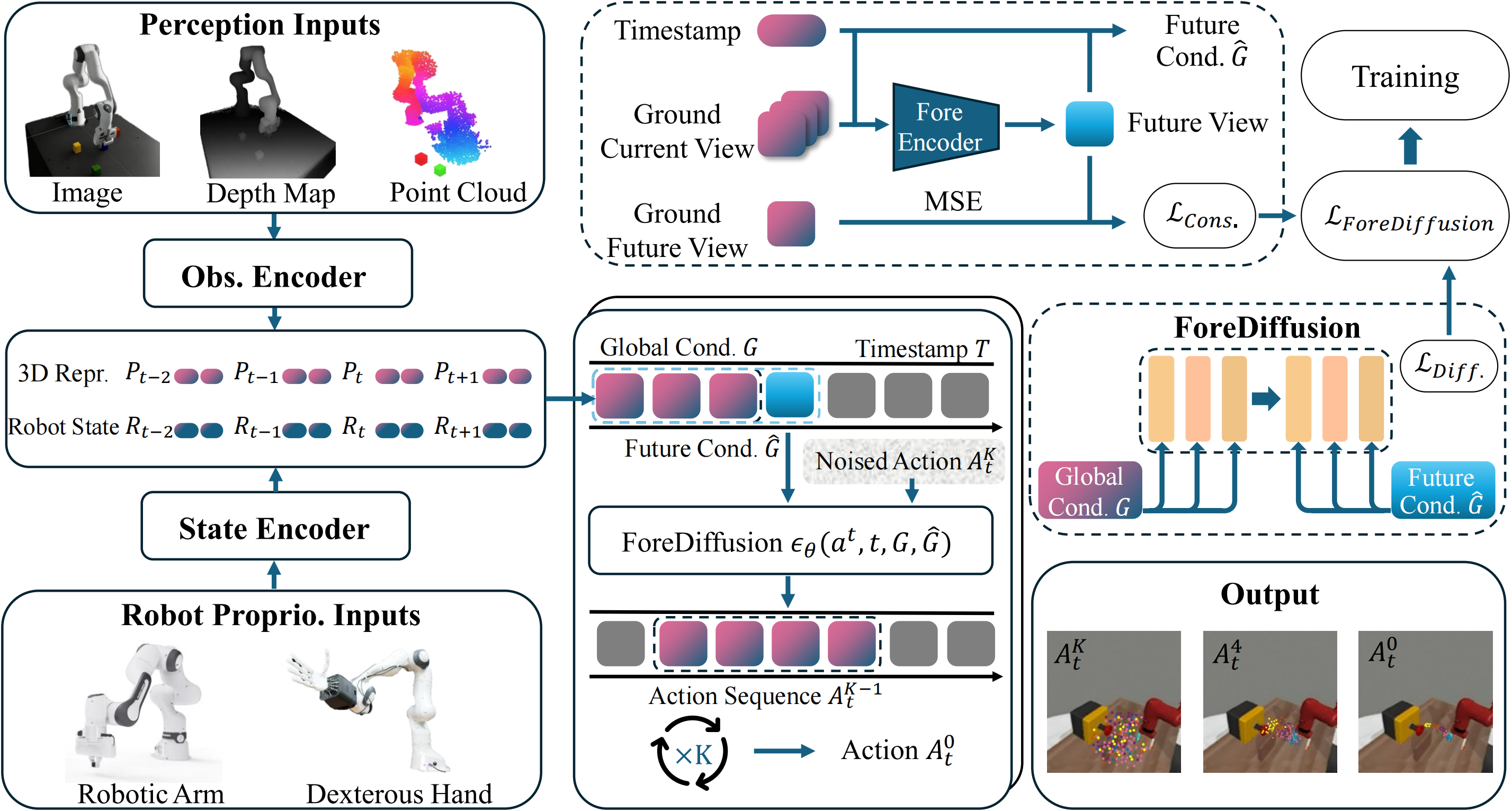}
    \caption{\textbf{Architecture of ForeDiffusion.} The perception module fuses RGB-D and proprioceptive inputs into 3D latent representations; an observation encoder outputs a global condition $G$ and a future condition $\hat{G}$. These condition guide a $K$-step reverse-diffusion process that denoises a noise-perturbed action trajectory into an executable sequence $A_0$, with joint construction and behavioral losses enforcing accurate future prediction and expert-level control.}
    \label{fig:Framework}
\end{figure*}

To stabilize training and reduce exposure bias, many works add task level consistency losses to the basic reconstruction or denoising objective~\cite{li2024stabilizing}. Representative examples include the self predictive loss in TD-MPC2~\cite{hansen2023td}, horizon wise KL regularization in Trajectory Transformer++~\cite{janner2021sequence}, and the score plus value distillation in SDMP~\cite{jia2024scoredistributionmatchingpolicy}. Collectively, prior work shows that combining explicit future prediction with joint local and global supervision leads to more reliable control on complex, long‑horizon tasks~\cite{chen2024simple}.

Building on 3D Diffusion Policy, to tackle the performance decline of complex tasks caused by limited perception and weak task supervision, we propose \textbf{ForeDiffusion}, a modular, foresight‑aware visuomotor policy. By conditioning each diffusion step on a compact, predicted future view and optimizing with a dual loss objective that balances construction fidelity and long‑horizon consistency, ForeDiffusion delivers more stable and anticipatory control. Experiments show that while it matches mainstream methods on basic tasks, it consistently surpasses them on complex, contact‑rich manipulation tasks, achieving markedly higher success rates under increasing task complexity.

\section{Method}
In this section, we first describe how to construct a future view from observations to approximate the future perception. Then, we develop how the current and future views are combined as conditions to guide the denoising process and generate trajectories that align with expert intent. Finally, we introduce a dual loss mechanism that jointly supervises both the future view construction and the denoising objectives.
\subsection{Future View Construction}
To enable the model to reason about future outcomes while relying on present information at inference, we introduce a future view construction mechanism that predicts a forward-looking representation from near-term observations. 
At each time step $t$, we define the current observation as a pair of temporally adjacent frames, $O^{cur.}_t = (O_{t-1}, O_t)$, where each $O_t = (P_t, R_t)$ consists of a 3D point cloud $P_t$ and proprioceptive robot state $R_t$~\cite{janner2022planning}. We encode this pair using a shared observation encoder $Enc(\cdot)$ to obtain the current observation $F^{cur.}_t$, defined as:
\begin{equation}
F^{cur.}_t = Enc\left( (P_{t-1}, R_{t-1}), (P_t, R_t) \right)
\end{equation}

The ground-truth future view \( F^{gt.}_t \) is constructed by the observation at time \( t+1 \). Based on the current observation feature \( F^{cur.}_t \), we construct a predictive future view \( F^{cons.}_t \) using a multilayer perceptron (MLP). The MLP learns to map the current observation to a future scene representation, enabling the model to predict future view based on near-term observations.

\begin{figure}[t]
    \centering
    \includegraphics[width=1.0\linewidth]{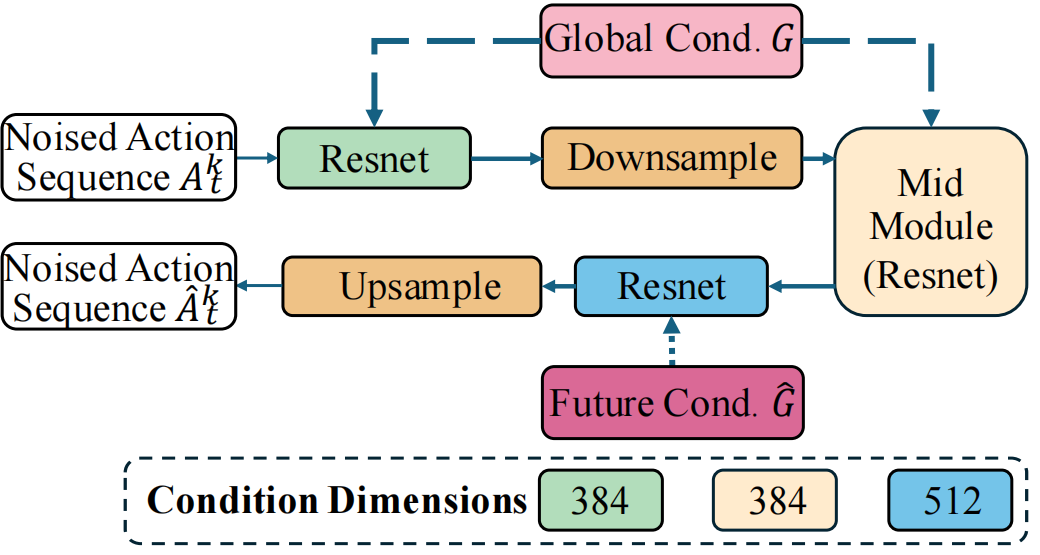}
    \caption{\textbf{Architecture of Foresight-Diffusion.} A ResNet encoder–decoder injects 384 / 512-dimensional context vectors \(G\) and \(\hat G\) across all sampling stages to denoise the action token \(A_{k}^{t}\) into \(\hat A_{k}^{t}\), with anticipated future observations.}
    \label{fig:Future_cond_construction}
\end{figure}

\subsection{Foresight-Conditioned Denoising}
The global condition $\mathbf{G}$ and the future condition $\mathbf{\hat{G}}$ denote separately the temporally-aware representations constructed by concatenating $F^{cur.}_t$ and $F^{cons.}_t$ with a timestamp encoding and enables the model to incorporate timing information into both the current and predicted future view. Our diffusion model generates actions by reversing a learned conditional diffusion process~\cite{ho2020denoising}. Starting from an initial Gaussian noise $\mathbf{a}^T \sim \mathcal{N}(0, I)$, the model iteratively refines the noisy action over $T$ denoising steps (see Fig.~\ref{fig:Future_cond_construction}). Each step is performed as follows:
\begin{multline}
    \mathbf{a}^{t-1} = \text{Denoise}(\alpha_t, \sigma_t,\mathbf{a}^t, t, \mathbf{G}), \\ = \alpha_t \left( \mathbf{a}^t - \frac{1 - \alpha_t}{\sqrt{1 - \bar{\alpha}_t}} \cdot \epsilon_\theta(\mathbf{a}^t, t, \mathbf{G}, \hat{\mathbf{G}}) \right) \\ + \sigma_t \cdot \mathbf{z}, \quad z \sim \mathcal{N}(0, I), \quad t = T, \dots, 1.
\end{multline}
where $\epsilon_\theta$ denotes a conditional denoising network that predicts the noise added to the action $\mathbf{a}^t$, given the timestep $t$, global condition $\mathbf{G}$, and future condition $\hat{\mathbf{G}}$. The denoising process is modulated by a variance schedule $\{\alpha_t, \bar{\alpha}_t, \sigma_t\}$, and the final output $\mathbf{a}^0$ serves as the predicted action.

This iterative procedure can be interpreted as a gradient-based update within an implicit energy field over the action space. Specifically, the denoising step resembles a descent in a learned energy field $E(x)$, yielding a simplified rule:

\begin{equation}
    \mathbf{a'} = \mathbf{a} - \gamma_t \cdot \nabla E(x)
\end{equation}
where $\gamma_t$ controls the step size and $\nabla E(x)$ represents the direction for refining the noisy action. While this abstraction omits scheduling details, it offers an intuitive view on how the diffusion model guide actions toward more optimal regions in the behavior space, shaped by current observations and predicted future view. 

\subsection{Dual Diffusion Loss}
To ensure that this constructed feature captures forward-looking information, we supervise it using a mean squared error loss against the encoded ground-truth future view:

\begin{equation}
    \mathcal{L}_{Construction} = \left\| F^{cons.}_t - F^{gt.}_t \right\|_2^2
\end{equation}

In parallel, the diffusion model is trained to reverse the noise process with a denoising objective:

\begin{equation}
    \mathcal{L}_{Diffusion} = E_{x_t, t} \left[ \left\| \epsilon_\theta(\mathbf{a}^t, t, \cdot) - \epsilon \right\|_2^2 \right]
\end{equation}

The total training objective is a weighted sum of the diffusion and structure alignment losses:

\begin{equation}
    \mathcal{L}_{ForeDiffusion} = \mathcal{L}_{Diff.} + \beta \cdot \mathcal{L}_{Cons.}
\end{equation}
where $\beta$ controls the influence of the prediction alignment. 

\begin{equation}
    \mathbf{a}^{t-1} = \mathbf{a}^t - \gamma_t \cdot \left( \epsilon_\theta^{Diff.}(\cdot) + \beta \cdot \epsilon_\theta^{Cons.}(\cdot) \right)
\end{equation}

This design enables the model to construct and utilize a structured predictive future view based solely on near-term observations, providing a form of implicit foresight that generalizes to inference without requiring access to future data.

\begin{figure}[t]
    \centering
    \includegraphics[width=1.0\linewidth]{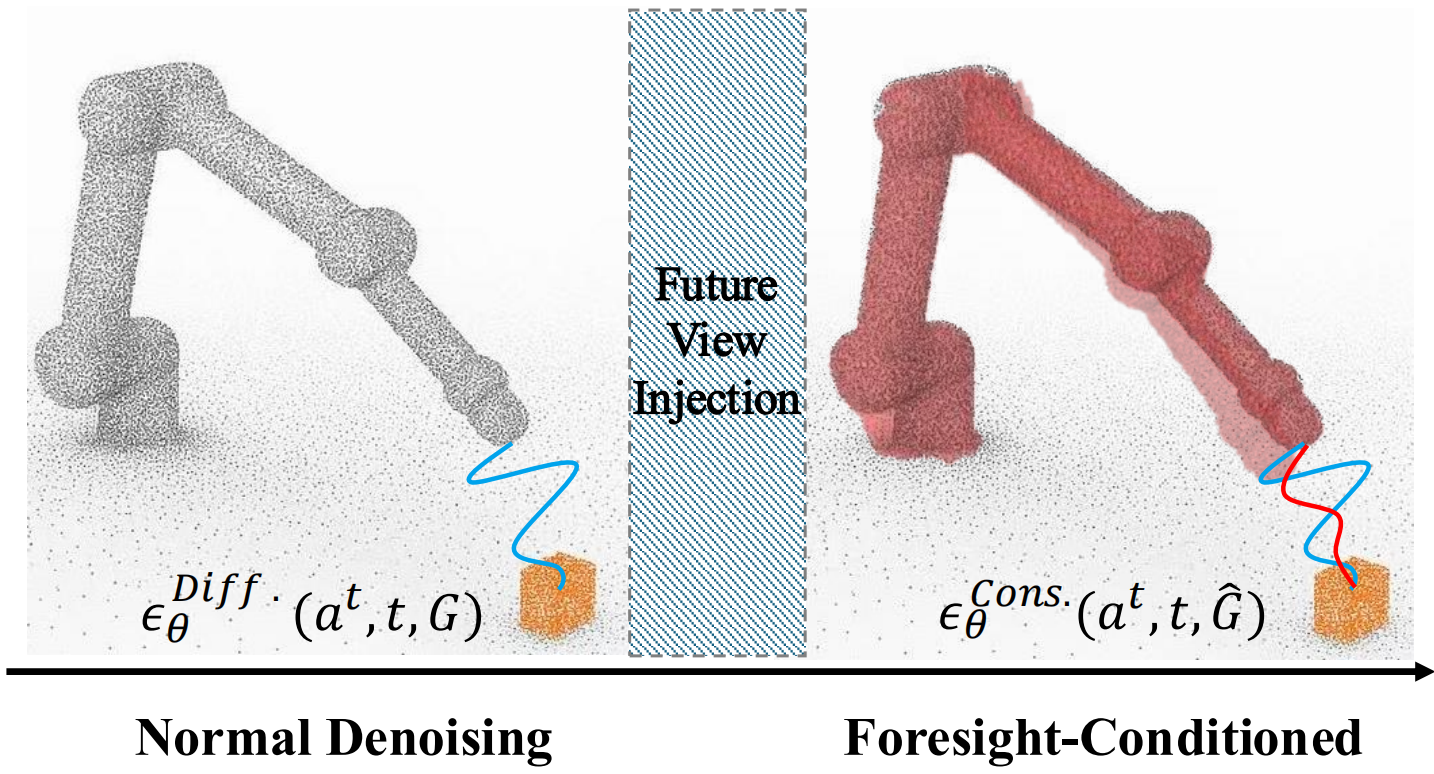}
    \caption{\textbf{Schematic of Future View Injection.} Normal denoising performs the standard early‑phase diffusion, whereas foresight conditioning injects future‑view information, allowing the model to anticipate consecutive outcomes and generate more stable, goal‑aligned action sequences.}
    \label{fig:Future_view_injection}
\end{figure}

In summary, at each time step $t$, ForeDiffusion extracts the current observation and predicts the future representation to form the diffusion conditions $(\mathbf{G}, \hat{\mathbf{G}})$ that guide each denoising step (see Fig.~\ref{fig:Future_view_injection}). Given expert trajectories, the goal is to learn a policy $\epsilon(a^t | O_{1:t})$ that mimics expert behavior. The training is supervised by a dual loss objective, combining the standard diffusion loss $\mathcal{L}_{Diff.}$ with an structure loss $\mathcal{L}_{Cons.}$ to improve both stability and fidelity.

\section{Intropy Analysis of ForeDiffusion}
The \textbf{Intropy framework} can be used to quantify intelligence as $d\mathcal{L} = \frac{\delta S}{R},$
where \( d\mathcal{L} \) (Intropy) denotes the incremental intelligence gain, \( \delta S \) represents newly absorbed information, and \( R \) reflects the system’s internal state, such as complexity or uncertainty \cite{RZY25}.

Under this view, \textbf{ForeDiffusion} maximizes \emph{intropy efficiency} by increasing information gain (\( \delta S \)) while reducing structural resistance (\( R \)). Its foresight-conditioned denoising, guided by predicted future views \( F^{\text{cons}}_t \), injects anticipatory information that amplifies long-horizon learning signals. The dual loss $\mathcal{L} = \mathcal{L}_{\text{Diff}} + \beta \cdot \mathcal{L}_{\text{Cons}},$
further stabilizes representation and reduces uncertainty. Through the Intropy lens, ForeDiffusion achieves greater \( d\mathcal{L} \)—transforming predictive entropy into structured foresight—yielding robust, intelligent visuomotor control with an average success rate of 80.6\% on long-horizon, contact-rich manipulation tasks.

\begin{table*}[t]
    \centering
    \begin{small}
    \begin{tabular}{
        l|
        >{\centering\arraybackslash}p{0.8cm}
        >{\centering\arraybackslash}p{0.8cm}
        >{\centering\arraybackslash}p{0.8cm}|
        >{\centering\arraybackslash}p{0.8cm}
        >{\centering\arraybackslash}p{0.8cm}
        >{\centering\arraybackslash}p{0.8cm}
        >{\centering\arraybackslash}p{1.4cm}|
        >{\centering\arraybackslash}p{3.2cm}}
    \toprule
    \multirow{2}{*}{Methods} & \multicolumn{3}{c}{Adroit Benchmark} & \multicolumn{4}{c|}{MetaWorld Benchmark} & \multirow{2}{*}{\textbf{Average}}\\
     & Hammer & Pen & Door & Easy & Medium & Hard & Very Hard & \\  \midrule
    DP~\cite{chi2023diffusionpolicy} & 45$\pm{5}$ & 13$\pm{2}$ & 37$\pm{2}$ & 
    82$\pm{26}$ & 31$\pm{24}$ & 11$\pm{13}$ & 37$\pm{27}$ & 56.64$\pm{37.61}$ \\
    FlowPolicy~\cite{zhang2025flowpolicy} & 100$\pm{0}$ & 53$\pm{12}$ & 58$\pm{5}$ & 80$\pm{34}$ & 62$\pm{32}$ & 49$\pm{40}$ & 36$\pm{7}$ & 62.57$\pm{21.00}$ \\
    ManiCM~\cite{lu2025manicmrealtime3ddiffusion} & 100$\pm{0}$ & 48$\pm{3}$ & \textbf{75$\pm{\textbf{3}}$} & 86$\pm{23}$ & 58$\pm{25}$ & 40$\pm{32}$ & 67$\pm{26}$ & 71.83$\pm{30.37}$ \\
    SDM Policy~\cite{jia2024scoredistributionmatchingpolicy} & 100$\pm{0}$ & 49$\pm{4}$ & 68$\pm{1}$ & 89$\pm{19}$ & 66$\pm{28}$ & 43$\pm{37}$ & 52$\pm{32}$ & 74.81$\pm{30.18}$ \\
    DP3~\cite{Ze2024DP3} & 100$\pm{0}$ & 43$\pm{6}$ & 62$\pm{4}$ & \textbf{91$\pm{\textbf{19}}$} & 61$\pm{32}$ & 44$\pm{41}$ & 49$\pm{33}$ & 72.35$\pm{32.77}$ \\
    \midrule
    \textbf{ForeDiffusion (Ours)} & \textbf{100$\pm{\textbf{0}}$} & \textbf{59$\pm{\textbf{12}}$} & 68$\pm{1}$ & 89$\pm{16}$ & \textbf{73$\pm{\textbf{26}}$} & \textbf{59$\pm{\textbf{29}}$} & \textbf{75$\pm{\textbf{28}}$} & \textbf{80.56$\pm{\textbf{22.93}}$} ($\uparrow$ 5.75\%) \\
       \bottomrule
    \end{tabular}
    \end{small}
    \caption{\textbf{Average success rates (\%) of ForeDiffusion and baselines on the Adroit and MetaWorld benchmarks.} ForeDiffusion achieves the top average success rate (80.56\%), matching or surpassing all baselines on every Adroit task and dominating the Medium, Hard, and Very Hard tiers of MetaWorld by margins of 6–26\%. Its consistently superior scores across the most challenging settings underscore the benefit of foresight-conditioned diffusion for robust grasping and long-horizon planning.}
    \label{tab:average_success_rate}
\end{table*}

\section{Experiments}
In this section, we conduct the experiments that focus on six research questions: \textbf{(RQ1)} Does the ForeDiffusion outperform baselines in task success rate? \textbf{(RQ2)} Is it effective for improving the learning efficiency across different tasks? \textbf{(RQ3)} As task complexity increases, does the performance remain consistently robust? \textbf{(RQ4)} Is foresight injection position essential for performance enhancement? \textbf{(RQ5)} How sensitive is the weighting of the dual loss during training? \textbf{(RQ6)} How does the scale of demonstrations influence performance and scalability?

\subsection{Experiment Setups}
\subsubsection{Simulation Benchmark} To evaluate our method, we adopt two representative simulation platforms as benchmarks: Adroit~\cite{rajeswaran2017learning} and MetaWorld~\cite{yu2020meta}. Adroit offers high-DoF dexterous hand manipulation with complex motor control; MetaWorld is implemented using the MuJoCo physics simulator~\cite{todorov2012mujoco}. It provides more robot operation tasks and serves as a standard benchmark for evaluating policy generalization and task transferability. Furthermore, as the length and complexity of the task time series increase, they are often categorized as Easy/Medium/Hard/Very Hard~\cite{mclean2025meta, seo2023masked}. These benchmarks cover the entire process from low-level control to high-level perceptual reasoning. To unify evaluation metrics, based on the MetaWorld classification criteria, we further termed the Medium, Hard, and Very Hard tiers of MetaWorld as complex tasks~\cite{hu2023imitation}.


\subsubsection{Expert Demonstrations} For expert demonstrations, we adopt domain-specific strategies to ensure high-quality and consistent supervision across tasks. In MetaWorld, expert trajectories are generated using built-in scripted policies. For other domains, expert data is collected from reinforcement learning agents trained to solve the tasks~\cite{nguyen2019review, schulman2017proximal}. Specifically, for Adroit, we employ VRL3~\cite{wang2022vrl3} to obtain successful trajectories. We ensure that all imitation learning algorithms are trained with the same set of expert trajectories. 

\subsubsection{Baselines} We compare against five representative diffusion based baselines. Diffusion Policy~\cite{chi2023diffusionpolicy} formulates single-stage conditional action generation. DP3~\cite{Ze2024DP3} extends this with 3D-conditioned modeling for visuomotor control. FlowPolicy~\cite{zhang2025flowpolicy} enforces trajectory coherence via implicit temporal modeling. ManiCM~\cite{lu2025manicmrealtime3ddiffusion} integrates multimodal inputs into a consistency-driven 3D diffusion framework.
SDM Policy~\cite{jia2024scoredistributionmatchingpolicy} adopts teacher–student distillation for accelerated inference.  These baselines span a spectrum of design philosophies from flat to structured diffusion, and from explicit to implicit behavior modeling.

\subsubsection{Implementation Details} We implement our method built upon a conditional U-Net backbone. FiLM-style~\cite{perez2018film} conditional modulation is applied throughout the network, with asymmetric conditioning dimensions: 384 for the downsampling and 512 for the upsampling. Point clouds are encoded via a PointNet-style encoder~\cite{qi2017pointnet} with 3 input channels and 64 output dimensions. We train the policy using AdamW with a learning rate of 1e-4, cosine learning rate schedule, and 500 warm-up steps. The model is trained for 3000 epochs with a batch size of 128. Diffusion is implemented with the DDIM scheduler~\cite{song2020denoising}, and EMA is applied to stabilize training. All experiments are performed on a single NVIDIA RTX 3080 GPU.

\subsubsection{Evaluation Metrics} Based on the evaluation protocol of DP3, each experiment is conducted with 3 random seeds (0, 1, 2). Throughout training, the policy is evaluated every 200 epochs using 15 rollouts over a total of 3,000 episodes. For each seed, we compute the average of the top 5 success rates across all evaluations, and report the final performance as the mean and standard deviation over the 3 seeds.

\subsection{Comparison with State-of-the-art Methods}
\begin{table*}[t]
    \centering
    \begin{small}
    \begin{tabular}{l|cccc|ccc|ccccc|c}
        \toprule
        \multirow{2}{*}{Methods} &
        \multicolumn{4}{c|}{Medium} &
        \multicolumn{3}{c|}{Hard} &
        \multicolumn{5}{c|}{Very Hard} &
        \multirow{2}{*}{\textbf{Average}} \\
        & B & BP & H & PW
        & HI & POH & P
        & D & PPW & SPe & SPl & SPh & \\ \midrule
        Diffusion Policy (DP)
        & 85 & 15 & 15 & 20
        & 9 & 0 & 30
        & 43 & 5 & 11 & 11 & 63
        & 26 \\
        3D Diffusion Policy (DP3)
        & 98 & 34 & 76 & 49
        & 14 & 14 & 51
        & 69 & 35 & 17 & 27 & 97
        & 48 \\ \midrule
        \textbf{ForeDiffusion (Ours)}
        & \textbf{100} & \textbf{39} & \textbf{97} & \textbf{99}
        & \textbf{20} & \textbf{53} & \textbf{75}
        & \textbf{94} & \textbf{92} & \textbf{39} & \textbf{49} & \textbf{100}
        & \textbf{71} ($\uparrow$ 23\%) \\
        \bottomrule
    \end{tabular}
    \end{small}
    \caption{\textbf{Success rates (\%) of different policies on challenging metaworld tasks.} ForeDiffusion delivers the highest score on every one of the complex tasks in MetaWorld, raising the overall mean success to 71\%, a gain of 23\% over DP3 and 45\% over the Diffusion Policy. The advantages are most evident on the Very Hard tasks, up to 92\% on PPW and 100\% on SPh. These results confirm that ForeDiffusion markedly improves robustness under increasing task complexity.}
    \label{tab:hard_task}
    \end{table*}
    
\begin{figure*}[t]
    \centering
    \includegraphics[width=1.0\linewidth]{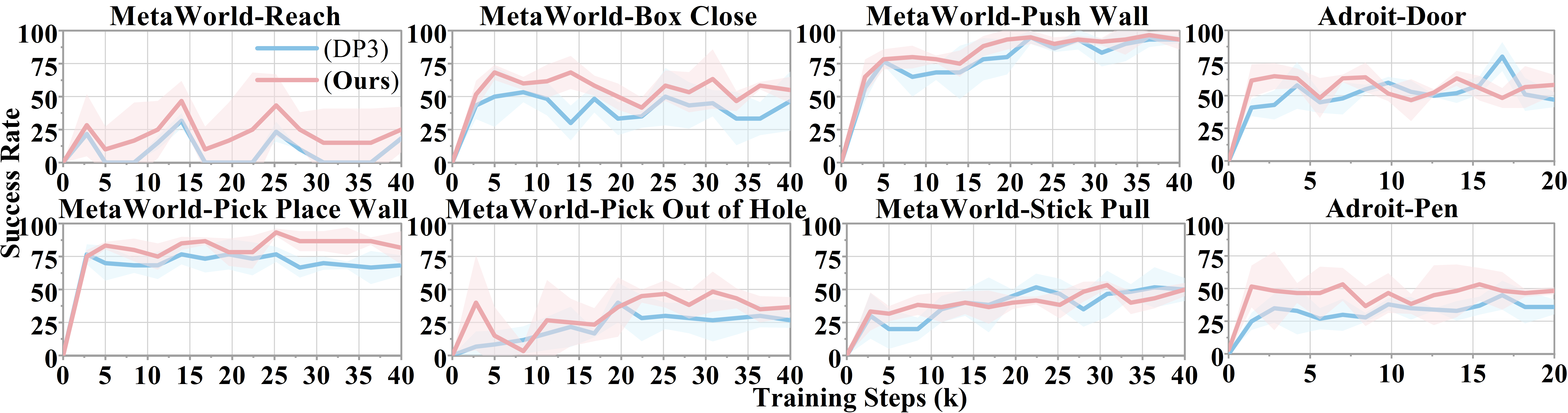}
    \caption{\textbf{Learning efficiency.} Compared to DP3, ForeDiffusion shows higher stability, learning efficiency, and success rates.}
    \label{fig:Learning_Efficiency}
\end{figure*}

\subsubsection{Success Rate (RQ1)} We compare the success rates of our method with state-of-the-art baselines on Adroit suits and MetaWorld benchmark. Our method consistently achieves competitive performance, particularly excelling in complex, dexterous and multi-task settings. Table~\ref{tab:average_success_rate} compares policy performance across the Adroit and MetaWorld benchmarks. On the Adroit tasks, ours achieves 100\% on Hammer, and also surpasses prior methods on Pen and Door. In MetaWorld, ForeDiffusion maintains strong results as task complexity increases. On Medium, Hard and Very Hard subsets, it reaches 73\%, 59\% and 75\% success rates, respectively, with standard deviations of 26, 29 and 28. These results represent gains of up to 26\% points over the best-performing baseline, and are particularly significant on Hard and Very Hard tasks where most baselines degrade sharply. FlowPolicy drops to 36\% on Very Hard, while DP3 achieves only 44\% on Hard. Overall, ForeDiffusion delivers an average success rate of 80.6\% with a standard deviation of 22.9 across all tasks, outperforming the strongest baseline SDM Policy (74.8\%).  Overall, the results underline ForeDiffusion's superior success rates and its robust, steady performance on the most demanding manipulation tasks.

\begin{figure}[t]
    \centering
    \includegraphics[width=1.0\linewidth]{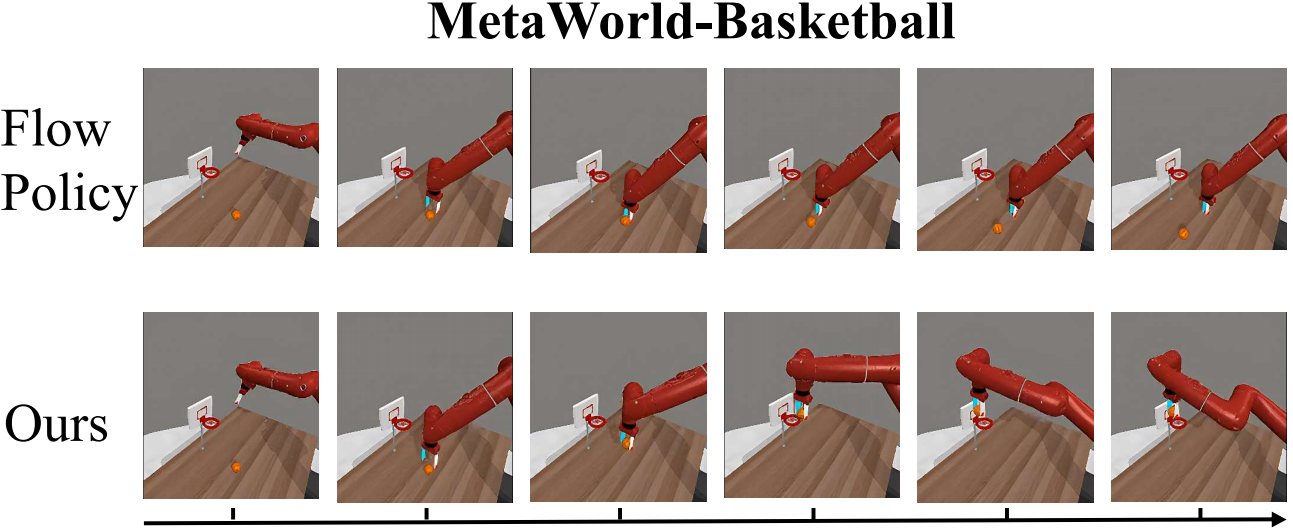}
    \caption{\textbf{Basketball task on different method.} ForeDiffusion consistently scores, whereas FlowPolicy misses, demonstrating superior grasp stability, trajectory planning.}
    \label{fig:MetaWorld-Basketball}
\end{figure}

\subsubsection{Learning Efficiency (RQ2)} To evaluate learning efficiency, we compare the success rate curves of different methods across tasks of varying complexity (see Fig.~\ref{fig:Learning_Efficiency}). Our method consistently outperforms DP3 under low-data regimes. On Disassemble, we achieve 95\% success with only 10 demonstrations, while DP3 reaches just 40\%. For harder tasks like Shelf Place, we reach 55\% with 10 demonstrations, whereas DP3 remains at 25\%. Even in tasks where DP3 eventually catches up (e.g. Stick Push), our method converges faster (100\% at 10 demos vs. 90\% for DP3 at 20). Overall, our policy achieves over 90\% success on 4 out of 5 tasks within 20 demonstrations, while DP3 requires 50 demonstrations to reach comparable performance.

\begin{figure*}[t]
    \centering
    \includegraphics[width=1.0\linewidth]{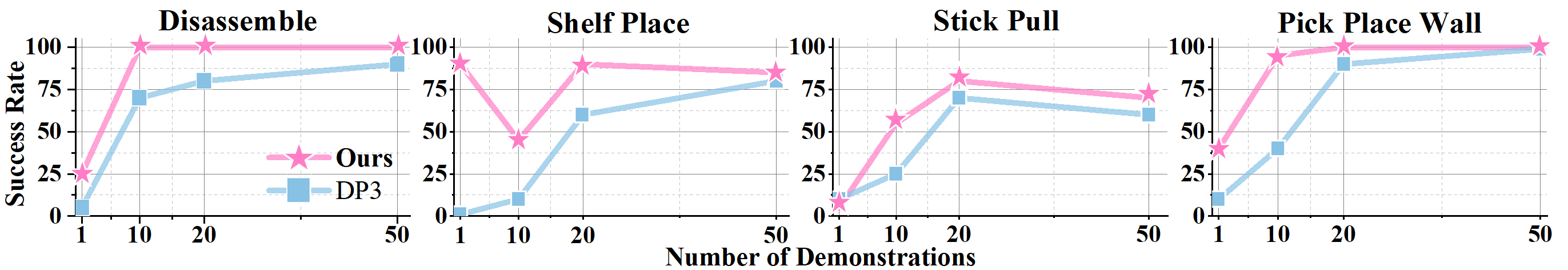}
    \caption{\textbf{Qualitative comparison of efficient scaling with demonstrations.} To evaluate the impact of demonstration quantity on policy performance, we conduct experiments on four Very Hard tasks from the MetaWorld. We compare ForeDiffusion and DP3 by progressively increasing the number of expert demonstrations during training. Results show that ForeDiffusion consistently achieves higher success rates than DP3 under low-data regimes, demonstrating strong sample efficiency.}
    \label{fig:Scaling_with_Expert_Demonstrations}
\end{figure*}

\subsubsection{Prospective Learning Ability (RQ3)} To evaluate the model adaptability and reasoning across rising task complexity, we evaluate Foresight‑Conditioned Diffusion on 12 MetaWorld tasks spanning the Medium, Hard, and Very Hard tasks in the MetaWorld benchmark: Basketball (\textbf{B}), Bin Picking (\textbf{BP}), Hammer (\textbf{H}), Push Wall (\textbf{PW}), Hand Insert (\textbf{HI}), Pick Out of Hole (\textbf{POH}), Push (\textbf{P}), Disasemble (\textbf{D}), Pick Place Wall (\textbf{PPW}), Shelf Place (\textbf{SPe}), Stick Pull (\textbf{SPl}), Stick Push (\textbf{SPh}). As shown in Table~\ref{tab:hard_task}, ours consistently surpasses all baselines, with the margin widening at complex tasks. This demonstrates the foresight module allows the policy to exploit future information, anticipate long‑horizon outcomes, and make more robust decisions capabilities that standard diffusion policies lose under high uncertainty or long temporal dependencies. Fig.~\ref{fig:MetaWorld-Basketball} compares FlowPolicy with our method on the same simulated task, confirming that ForeDiffusion excels on challenges demanding long‑term reasoning and planning. 
   
\subsection{Ablations}
\subsubsection{Foresight-Conditioned Diffusion (RQ4)} To evaluate the effectiveness of our method, we conduct ablation studies focusing on the point at which future view is injected into the diffusion process. In the baseline, future view is entirely removed, and the model relies solely on current view, allowing us to assess the contribution of foresight-conditioned diffusion itself. In the early-stage, future view are fused at the input stage of the U-Net alongside global conditioning, influencing the entire diffusion process from the beginning. In contrast, ForeDiffusion introduces future view at the mid-stage of the U-Net, directly altering the downstream generative trajectory while preserving the early-stage representation learning (see Table~\ref{tab:ablation_foresight}). Experimental results show that removing foresight leads to significantly worse performance and reduced policy stability, while early injection offers limited gains due to the dilution of future context. Our mid-stage injection achieves the best performance, validating that conditioning diffusion at structurally critical stages enables more effective use of future view.

\begin{table}[t]
    \centering
    \begin{small}
    \setlength{\tabcolsep}{1mm}
    \begin{tabular}{l|ccc|c}
    \toprule
    \multirow{2}{*}{Injection Position} & \multicolumn{3}{c|}{MetaWorld Benchmark} & \multirow{2}{*}{\textbf{Average}}\\
     & Medium & Hard & Very Hard & \\  \midrule
    w/o Future view & 61$\pm{32}$ & 44$\pm{41}$ & 49$\pm{33}$ & 50.5$\pm{32.6}$ \\
    \midrule
    \textbf{Early (Ours)} & 71$\pm{29}$ & \textbf{60$\pm{\textbf{29}}$} & 66$\pm{31}$ & 67.7$\pm{28.1}$ \\
    \textbf{Mid-Stage (Ours)} & \textbf{73$\pm{\textbf{26}}$} & 59$\pm{29}$ & \textbf{75$\pm{\textbf{28}}$} & \textbf{70.3$\pm{\textbf{27.3}}$} \\ \bottomrule
    \end{tabular}
    \end{small}
    \caption{\textbf{Ablation results on the MetaWorld compare different future view injection position.} Introducing foresight from the mid-stage of the diffusion leads to the best overall performance across Medium and Very Hard tasks.}
    \label{tab:ablation_foresight}
\end{table}

\begin{table}[t]
    \centering
    \begin{small}
    \setlength{\tabcolsep}{1mm}
    \begin{tabular}{l|ccc|c}
    \toprule
    \multirow{2}{*}{Dual Loss Weight} & \multicolumn{3}{c|}{MetaWorld Benchmark} & \multirow{2}{*}{\textbf{Average}}\\
     & Medium & Hard & Very Hard & \\  \midrule
    w/o Dual Loss & 71$\pm{29}$ & 58$\pm{32}$ & 66$\pm{30}$ & 67.8$\pm{28.6}$ \\
    \midrule
    \textbf{Dynamic (Ours)} & 70$\pm{30}$ & \textbf{62$\pm{\textbf{31}}$} & 69$\pm{30}$ & 68.3$\pm{28.7}$ \\
    \textbf{Fixed (Ours)} & \textbf{73$\pm{\textbf{26}}$} & 59$\pm{29}$ & \textbf{75$\pm{\textbf{28}}$} & \textbf{70.3$\pm{\textbf{27.3}}$} \\ \bottomrule
    \end{tabular}
    \end{small}
    \caption{\textbf{Ablation studies on the effect of dual-loss strategies under different difficulty levels in MetaWorld.} The fixed weight dual loss achieves the best performance on Medium and Very Hard tasks and the highest average score.}
    \label{tab:ablation_dual_loss}
    \end{table}
    
\subsubsection{Dual Loss Synergy (RQ5)} In order to verify the effectiveness of the proposed dual loss mechanism, we design three groups of comparative experiments: the first group completely removes the auxiliary loss and only use the diffusion prediction loss for training; the second group adopts a dynamic weight strategy, in which the proportion of the auxiliary loss increases with training steps, following an exponential growth, allowing the model to progressively adjust its reliance on supervision signals at different stages; the third group is the main method of this paper, which uses fixed weights to fuse the main loss and auxiliary loss. To evaluate the impact of the dual loss design, we perform ablation studies on the MetaWorld benchmark under varying task complexities. As shown in Table~\ref{tab:ablation_dual_loss}, removing the dual loss significantly degrades performance, especially on Very Hard tasks, suggesting that dual loss plays a critical role in guiding the diffusion process. Although dynamically weighting the two losses improves results, it still suffers from instability across tasks. In contrast, our fixed weighting dual loss formulation achieves the best overall performance, reaching an average success rate of 70.3\%. This highlights the effectiveness and stability of incorporating future view supervision in a balanced and consistent manner.

\subsubsection{Scaling with Expert Demonstrations (RQ6)} We evaluate the scalability of different methods with respect to the number of expert demonstrations, focusing on four Very Hard tasks from MetaWorld. As shown in Fig.~\ref{fig:Scaling_with_Expert_Demonstrations}, our method consistently outperforms DP3 across all demonstration settings. Remarkably, even with as few as a single demonstration, our method achieves strong performance, reaching success rates above 75\% on some tasks. This indicates that our method is able to extract and generalize task-relevant information efficiently, owing to the inductive structure imposed by foresight representation learning. As the number of demonstrations increases, our performance continues to improve steadily, widening the gap with baseline methods.

\section{Conclusion}
In this work, we propose ForeDiffusion, a foresight conditioned diffusion policy that predicts future view and optimizes long-horizon consistency, overcoming the short-horizon conditioning and single loss constraints of existing approaches, thereby preventing the sharp drop in success rate that occurs as task complexity rises.
A compact future-view latent is injected at the diffusion process, and a dual-loss design balances denoising fidelity with trajectory-level coherence to keep performance stable over extended horizons.
On the Adroit and MetaWorld benchmarks, it achieves an average success rate of 80.6\% and surpasses 3D Diffusion Policy by up to 23\% in complex manipulation tasks, demonstrating strong generalization, data efficiency, and interpretability.
We hope that ForeDiffusion drives further exploration to improve the performance of diffusion strategies towards manipulation applications.

\section{Acknowledgments}
This work is supported in part by Shenzhen Science and Technology Program under Grant ZDSYS20220527171400002, the National Natural Science Foundation of China (NSFC) under Grants 62271324, 62231020 and 62371309.
\bibliography{aaai2026}

\end{document}